%% file: main.tex
\documentclass[conference]{IEEEtran}
\IEEEoverridecommandlockouts
\usepackage{cite}
\usepackage{amsmath,amssymb,amsfonts}
\usepackage{mathtools}
\usepackage{algorithmic}
\usepackage{graphicx}
\usepackage{textcomp}
\usepackage{xcolor}
\usepackage{lipsum}
\usepackage{tabularray}
\def\BibTeX{{\rm B\kern-.05em{\sc i\kern-.025em b}\kern-.08em
    T\kern-.1667em\lower.7ex\hbox{E}\kern-.125emX}}
\begin{document}
\hyphenation{mo-del-s}
\hyphenation{met-ho-do-lo-gy}

\title{{One Model to Forecast Them All \\ and in Entity Distributions Bind Them}
\thanks{This research was undertaken as part of the InnoCyPES project, which has received funding from the European Union's Horizon 2020 research and innovation programme under the Marie
Sk{\l}odowska-Curie grant agreement No 956433.}
}

\author{\IEEEauthorblockN{Kutay Bölat}
\IEEEauthorblockA{\textit{Electrical Sustainable Energy} \\
\textit{Delft University of Technology}\\
Delft, Netherlands \\
K.Bolat@tudelft.nl}
\and
\IEEEauthorblockN{Simon H. Tindemans}
\IEEEauthorblockA{\textit{Electrical Sustainable Energy} \\
\textit{Delft University of Technology}\\
Delft, Netherlands \\
S.H.Tindemans@tudelft.nl}
}

\maketitle

\begin{abstract}
Probabilistic forecasting in power systems often involves multi-entity datasets like households, feeders, and wind turbines, where generating reliable entity-specific forecasts presents significant challenges. Traditional approaches require training individual models for each entity, making them inefficient and hard to scale. This study addresses this problem using GUIDE-VAE, a conditional variational autoencoder that allows entity-specific probabilistic forecasting using a single model. GUIDE-VAE provides flexible outputs, ranging from interpretable point estimates to full probability distributions, thanks to its advanced covariance composition structure. These distributions capture uncertainty and temporal dependencies, offering richer insights than traditional methods. 

To evaluate our GUIDE-VAE-based forecaster, we use household electricity consumption data as a case study due to its multi-entity and highly stochastic nature. Experimental results demonstrate that GUIDE-VAE outperforms conventional quantile regression techniques across key metrics while ensuring scalability and versatility. These features make GUIDE-VAE a powerful and generalizable tool for probabilistic forecasting tasks, with potential applications beyond household electricity consumption.
\end{abstract}


\input{chapters/introduction}
\input{chapters/setting}
\input{chapters/methodology}

\input{chapters/experiments}
\input{chapters/results}

\input{chapters/conclusion}

\bibliographystyle{IEEEtran}
\bibliography{references}

\end{document}

%% file: chapters/introduction.tex
\section{Introduction}

Uncertainty is inherent in modern power and energy systems, intensified by the growing integration of renewable energy sources and demand response programs that require close monitoring of distribution grids. Reliable forecasting is critical for ensuring system reliability, efficient resource allocation, and operational affordability in these volatile environments. While point forecasting methods provide deterministic estimates, they fail to capture uncertainty. Probabilistic forecasting addresses this limitation by offering richer insights through probability distributions, confidence intervals, or quantiles, enabling better risk management and planning \cite{wang2024generative}.

\begin{figure}
    \centering
    \includegraphics[width=1.0\linewidth]{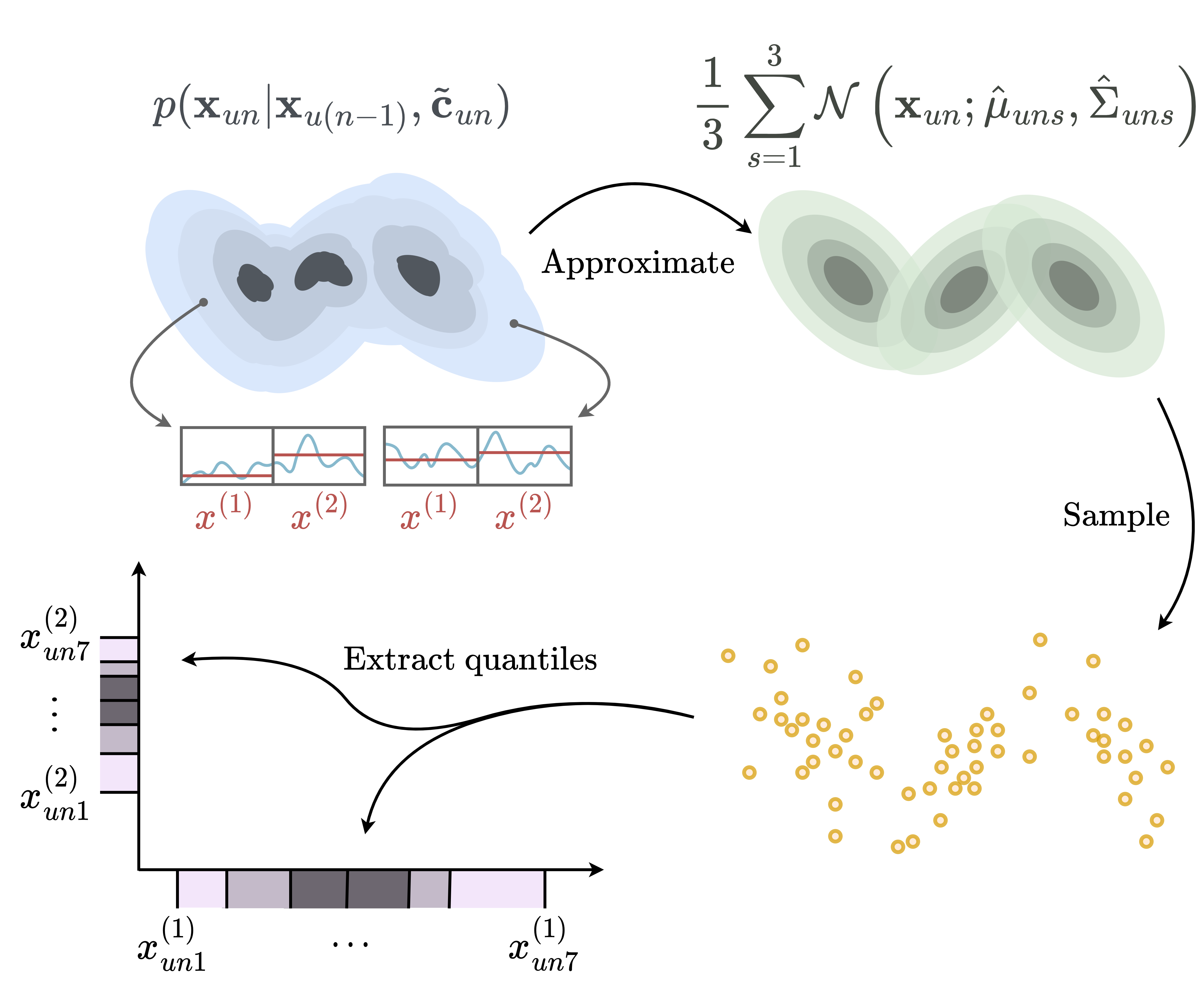}
    \caption{The different utilization levels of probabilistic forecasting are illustrated in a two-dimensional space, representing average load in subsequent 12-hour intervals. First, the (possibly) complex forecasting distribution is approximated as a mixture of Gaussians. Then, a collection of samples is taken from it to form an ensemble. Finally, quantiles of each marginal are extracted using this ensemble.}
    \label{fig:representations}
\end{figure}

Forecasting in power systems often involves multi-entity datasets, such as households, feeders, and wind turbines. Generating reliable forecasts for individual entities in such datasets poses significant challenges due to their inherent variability and complex dependencies. Traditional approaches typically require training separate models for each entity, leading to inefficiencies in computational resources and scalability. Furthermore, widely used methods such as quantile regression focus on marginal behaviour and struggle to model multivariate dependencies, particularly in day-ahead forecasting tasks where temporal correlations play a critical role \cite{haben2021review}. As depicted in Fig. \ref{fig:representations}, quantile regression fails to capture the joint dependency structure, limiting its effectiveness in applications requiring a full probabilistic distribution.

Recent advancements in deep generative models, such as Variational Autoencoders (VAEs) \cite{kingma2013auto} and normalizing flows \cite{rezende2015variational}, have provided powerful tools for modelling complex probability distributions. These methods capture multivariate uncertainties by learning the underlying patterns in high-dimensional datasets \cite{dumas2022deep}. However, their direct application to multi-entity datasets in power systems remains underexplored.

We adopt GUIDE-VAE \cite{bolat2024guide}, a conditional Variational Autoencoder designed for multi-entity and multivariate data, to address these challenges in modelling. GUIDE-VAE combines deep generative modelling with two advanced features: Pattern Dictionary-based Covariance Composition (PDCC) and probabilistic entity embeddings. These features enable GUIDE-VAE to perform parallel probabilistic forecasting for individual entities using a single model, eliminating the inefficiency of training separate models. GUIDE-VAE also provides flexible forecasting outputs, from interpretable point estimates to full probability distributions, capturing uncertainty and temporal dependencies \cite{jrhilifa2024forecasting}.

In this study, we focus on household electricity consumption as a challenging case study for multi-entity probabilistic forecasting. Household data is highly volatile and stochastic, making it an ideal testbed to demonstrate GUIDE-VAE’s scalability and modelling capabilities. The key contributions of this work are: (1) adapting GUIDE-VAE for day-ahead forecasting of multi-entity electricity consumption, (2) providing forecast outputs as distributions, samples, and quantiles, and (3) validating its effectiveness through experiments using household electricity consumption data.

\textbf{\textit{Notation.}} Vectors are bolded, and matrices are capitalized and bolded. The $\sim$ operator represents sampling, and a hat indicates the variable is a product of a sampling process. Depending on their contents, square brackets are used either for integer listing, $[S]=\{1,2, \dots, S\}$, for concatenation, $\mathbf{a}=[a^{(i)}]_i$, or for vector representation $\mathbf{v}=[a,b,c]$.

%% file: chapters/setting.tex
\section{Problem Setting}

We consider the problem of day-ahead probabilistic forecasting of individual entities' load profiles. Specifically, we aim to model the conditional probability distribution function (pdf) $  p(\mathbf{x}_{un}|\mathbf{x}_{u(n-1)},\tilde{\mathbf{c}}_{un})$ 
where $ \mathbf{x}_{un} \in \mathbb{R}^T $ represents the daily load profile of the $ u $-th entity on the $ n $-th day, and $ \tilde{\mathbf{c}}_{un} \in \mathbb{R}^{\tilde{C}} $ denotes contextual information, such as day-specific and entity-specific features.\footnote{The look-back window \( \mathbf{x}_{u(n-1)} \) is not restricted to the ``day-before" and can be extended as needed.}

The forecasting distribution is multivariate, meaning it captures the interdependencies between time steps: $p(\mathbf{x}_{un}|\mathbf{x}_{u(n-1)},\tilde{\mathbf{c}}_{un}) \neq \prod_t p(x_{un}^{(t)}|\mathbf{x}_{u(n-1)},\tilde{\mathbf{c}}_{un})$. To model this distribution, we employ GUIDE-VAE, an advanced VAE-based framework that effectively captures temporal and multivariate dependencies, as well as entity-level variability.

Since GUIDE-VAE does not yield an explicit pdf after training, we approximate the forecasting distribution using Monte Carlo sampling with $S$ elementary distributions:
\begin{equation}\label{eq:forecasting_pdf}
    \begin{split}
    p(\mathbf{x}_{un}|\mathbf{x}_{u(n-1)},\tilde{\mathbf{c}}_{un}) &= \mathbb{E}_{p(\mathbf{z})}[p(\mathbf{x}_{un}|\mathbf{x}_{u(n-1)},\tilde{\mathbf{c}}_{un}, \mathbf{z})] \\ &\approx \frac{1}{S} \sum_{s=1}^S p(\mathbf{x}_{un}|\mathbf{x}_{u(n-1)},\tilde{\mathbf{c}}_{un}, \hat{\mathbf{z}}_s) \\
    &= \frac{1}{S} \sum_{s=1}^S \mathcal{N}\left(\mathbf{x}_{un}; \hat{\mu}_{uns}, \hat{\Sigma}_{uns}\right)
    \end{split}
\end{equation}
where $\hat{\mathbf{z}}_s\sim p(\mathbf{z}), \forall s\in[S]$ are latent space samples and $\mathcal{N}(.;\mu,\Sigma)$ represents the multivariate normal distribution with mean vector $\mu$ and covariance matrix $\Sigma$. This approximation is depicted in Fig. \ref{fig:representations}.

This mixture of Gaussians (MoG) composition has several benefits. Firstly, the non-Gaussian nature of the mixture makes it highly flexible and elevates the modelling power. Secondly, it gives a tractable likelihood calculation even though the forecasting distribution does not have a closed-form representation. Lastly, it provides a simple sampling mechanism by two-level sampling, i.e. $p(\mathbf{z})\overset{\sim}{\rightarrow}\hat{\mathbf{z}} \rightarrow \mathcal{N}(\mathbf{x}_{un}; \hat{\mu}_{un}, \hat{\Sigma}_{un}) \overset{\sim}{\rightarrow} \hat{\mathbf{x}}_{un}$.

Operationally, this forecasting methodology can be utilized in three ways:
\begin{enumerate}
    \item \textbf{Likelihood assessment}: The forecasting pdf evaluates the likelihood of given point forecasts.
    \item \textbf{Scenario generation}: Samples from the pdf serve as ensembles of point forecasts.
    \item \textbf{Uncertainty quantification}: Quantiles extracted from the ensemble provide interpretable bounds for each time step.
\end{enumerate}
All of these possible utilizations are depicted in Fig. \ref{fig:representations}. It is important to note that with each ``simplification” step, the forecast becomes more interpretable at the cost of information loss. For example, quantiles ignore interdependencies among time steps, sacrificing the multivariate structure.

%% file: chapters/methodology.tex
\section{Methodology}

\begin{figure*}
    \centering
    \includegraphics[width=1.0\linewidth]{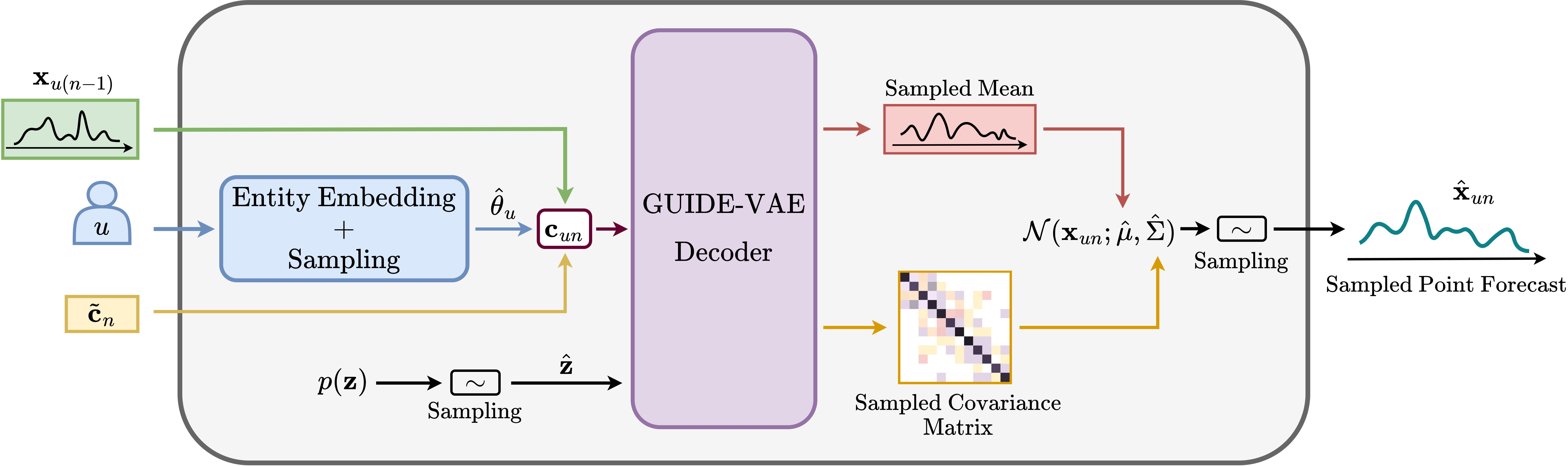}
    \caption{Overall diagram of GUIDE-VAE forecaster, which accepts the day-before consumption profile $\mathbf{x}_{u(n-1)}$, entity index $u$ and the contextual information vector $\tilde{\mathbf{c}}_n$ regarding the day $n$. Only one MoG component's generation is depicted here ($S=1$). In order to approximate the forecasting distribution in \eqref{eq:forecasting_pdf}, this sampling procedure is repeated by taking more samples from the prior $p(\mathbf{z})$, and generated mean vectors and covariance matrices are stored. Similarly, resulting point forecasts ($\hat{\mathbf{x}}_{un}$) sampled from these mixture components are stored to form the ensemble and, consequently, the quantile forecasts.} 
    \label{fig:forecast_diagram}
\end{figure*}

GUIDE-VAE \cite{bolat2024guide} is a conditional VAE model that approximates the probability distribution model\footnote{In order to optimize such a distribution model, all VAE-based models, including GUIDE-VAE, employ an amortized variational inference methodology. This requires an extra distribution model for the approximate posterior (a decoder network), which maps observable and conditioning variables into latent variables. Since this posterior distribution is essential only for the training of GUIDE-VAE and not significant for the proposed methodology, we refer interested readers to \cite{kingma2013auto} and \cite{bolat2024guide} for further details.} $p_{\psi,\mathbf{U}}(\mathbf{x}|\mathbf{c})=\mathbb{E}_{p(\mathbf{z})}\left[ p_{\psi,\mathbf{U}}(\mathbf{x}|\mathbf{z},\mathbf{c}) \right]$. Here, $
\mathbf{x}\in\mathbb{R}^T$, $\mathbf{c}\in\mathbb{R}^C$ and $\mathbf{z}\in\mathbb{R}^Z$ represent observable, conditioning and latent variables, respectively. This structure allows generating data points for a given condition $\mathbf{c}$ using ancestral sampling as $\hat{\mathbf{x}} \sim p_{\psi,\mathbf{U}}(\mathbf{x}|\hat{\mathbf{z}},\mathbf{c})$ where $\hat{\mathbf{z}} \sim p(\mathbf{z})$ as illustrated in Fig. \ref{fig:forecast_diagram}. Here, the likelihood and prior distributions are respectively modelled as
\begin{align}
        p_{\psi,\mathbf{U}}(\mathbf{x}|\mathbf{z},\mathbf{c}) &= \mathcal{N}\left(\mathbf{x}; \mu = \mu_\psi(\mathbf{z},\mathbf{c}), \Sigma=\Sigma_{\psi,\mathbf{U}}(\mathbf{z},\mathbf{c})\right) \label{eq:guide_vae_likelihood} \\
        p(\mathbf{z}) &= \mathcal{N}\left(\mathbf{z}; \mu=\mathbf{0}, \Sigma=\mathbf{I}\right).
\end{align}
The parameterization function for the mean $\mu_\psi$ can be chosen without a constraint and often modelled using a neural network $f^\mu_\psi(\mathbf{z},\mathbf{c})$. However, the parameterization of the covariance requires satisfying positive definiteness, and common practice is to model it as a positive diagonal matrix. Unfortunately, this choice results in an independence between the components of generated samples and hinders the modelling power.

Instead, GUIDE-VAE introduces PDCC, which enables the construction of any covariance matrix in a computationally efficient way. For this purpose, PDCC utilizes a large pattern dictionary $\mathbf{U}\in\mathbb{R}^{T\times V}$ with $V>T$, which stores representative patterns observed across the dataset, allowing GUIDE-VAE to model complex correlations between time steps. It is integrated into the covariance parameterization of the likelihood distribution in \eqref{eq:guide_vae_likelihood} as
\begin{equation}
    \Sigma_{\psi,\mathbf{U}}(\mathbf{z},\mathbf{c}) = \mathbf{U}\text{diag}(f^{\tilde{\sigma}}_\psi(\mathbf{z},\mathbf{c}))^2\mathbf{U}^\top + \xi\mathbf{I}
\end{equation}
where $f^{\tilde{\sigma}}_\psi:\mathbb{R}^{Z+C}\rightarrow\mathbb{R}_+^{V}$ is a neural network for the high dimensional auxiliary standard deviations, and $\xi>0$ is a hyperparameter to maintain numeric stability. Note that $\mathbf{U}$ does not depend on the inputs, yet it is still learned. This results in a global pattern dictionary storing representative patterns commonly occurring in data and increasing the modelling power and realism of the generated data points thanks to the captured correlations.

Another feature of GUIDE-VAE is the ability to incorporate entity information when it is used for modelling multi-entity datasets like smart meter measurements collected from a set of customers. This incorporation is held by applying the user information to the network as a conditioning variable, which provides additional modelling power and controlled entity-specific data generation. For this purpose, an entity embedding pipeline similar to \cite{chen2022constructing} is applied to the data records of individual entities and each of them is embedded in a vector $\gamma_u\in\mathbb{R}_+^K,~ \forall u \in [U]$ which represents the concentration parameter of a Dirichlet distribution that models the entity $u$ out of $U$ entities. In \cite{bolat2024guide}, samples from the corresponding Dirichlet distribution $\hat{\theta}_u\sim\text{Dir}(\gamma_u)$ are used to capture the uncertainty among the entities, and this study follows the same methodology.

Now that the fundamentals of the proposed forecasting pdf are set, we can adapt the GUIDE-VAE to the probabilistic forecasting setting. For this, first, let us represent our multi-entity dataset as $\mathcal{X}=\bigcup_{u=1}^U\mathcal{X}_u$ which consists of $U$ entity datasets $\mathcal{X}_u$. Each entity dataset contains a collection of $N$ daily time-series profiles $\mathbf{x}_{un}\in\mathbb{R}^T$, i.e. $\mathcal{X}_u=\{ \mathbf{x}_{un} \}_{n=1}^N$. We utilize GUIDE-VAE to model \eqref{eq:forecasting_pdf} by (i) integrating the entity vector $\hat{\theta}_u$ as a contextual information regarding the entity, i.e. $\tilde{\mathbf{c}}_{un} = [\hat{\theta}_u, \tilde{\mathbf{c}}_{n}]$, and (ii) interpreting the ``look-back" profile $\mathbf{x}_{u(n-1)}$ and the contextual information $\tilde{\mathbf{c}}_{un}$ as conditions for the likelihood function. Therefore, the conditioning variable in \eqref{eq:guide_vae_likelihood} becomes 
\begin{equation}
    \mathbf{c}_{un} = \left[\mathbf{x}_{u(n-1)}, \hat{\theta}_u, \tilde{\mathbf{c}}_{n} \right],
\end{equation}
where $\tilde{\mathbf{c}}_{n}$ is the vector containing the contextual information about the timestamp $n$ such as months and weekdays. The resulting GUIDE-VAE forecaster is depicted in Fig. \ref{fig:forecast_diagram}. Note that GUIDE-VAE forms an all-in-one forecaster that can provide all the utilization levels depicted in Fig. \ref{fig:representations}. This is due to its direct objective of modelling the forecasting distribution itself, unlike the models trained to predict quantiles or generate ensembles for forecasting.

%% file: chapters/experiments.tex
\section{Experiments}

\subsection{Dataset}
 This study uses a dataset of smart meter data collected from electricity consumers across 47 provinces in Spain, including homes, offices, and businesses \cite{quesada2024electricity}. The dataset spans from November 2014 to June 2022 and includes hourly electricity consumption records for 25,559 users. For the experiments, a subset was extracted focusing on data from Gipuzkoa, the province with the highest data density, between June 2021 and June 2022. Only users with at least one full year of continuous enrollment during this period were included, while those with negative and consistently zero consumption were excluded. The final dataset comprises $U$=6,830 users, each with $N$=365 daily profiles (24 hourly values per day, $T$=24), amounting to approximately 2.5 million records.

 Since the data is zero-inflated, we used the same hyperparameters to apply a zero-preserved log-normalization to each feature as described in \cite{bolat2024guide}. After normalization, we reserved the last 73 days of all users' records (20\% of the whole dataset) as the testing set. At the same time, the remaining portion was split into training and validation sets with a ratio of 3:1. We used the validation set for early stopping and learning rate scheduling of every model we trained.

 We conduct the experiments using two auxiliary conditioning features: months and weekdays. These conditions are derived from the data index \( n \), transformed using a cyclic (\( \sin \)-\( \cos \)) encoding as proposed in \cite{wang2022contextual}, and associated with their respective data points \( \mathbf{x}_{un} \). As for the user embeddings, we used the same hyperparameter setting employed in \cite{bolat2024guide}.

\subsection{Baseline Model}
We benchmarked our methodology using a multi-head quantile regression neural network as the baseline model. For a given set of ordered quantile levels $\mathbf{q}=[q^{(i)}]_{i=1}^Q \in [0,1]^Q: q^{(i)}<q^{(i+1)}~ \forall i\in[Q] $, the regression model is defined as
\begin{equation}
    \begin{split}
    y^{(t)}_0 &= f_{\phi}^{(t,0)}(\mathbf{x},\mathbf{c}), ~\forall t\\
    y^{(t)}_i &= y^{(t)}_0 + \sum_{j=1}^i f_{\phi}^{(t,j)}(\mathbf{x},\mathbf{c}), ~ \forall i\geq1, \forall t
    \end{split}
\end{equation}
where $y^{(t)}_i$ represents the $q^{(i)}$-th quantile value of the $t$-th feature for a given regressor $\mathbf{x}$ and contextual information $\mathbf{c}$ mapped using a neural network $f_\phi: \mathbb{R}^{T+C}\rightarrow\mathbb{R}^{T\times Q}$ with the constraint $f_{\phi}^{(t,i)}>0, \forall i\geq1, \forall t$. This model can be trained by minimizing the following loss function
\begin{equation}\label{eq:quantile_loss}
\begin{split}
    \mathcal{L} = \frac{1}{MQT} \sum_{m,i,t}^{M,Q,T} &\max\left((q^{(i)}-1)\varepsilon_{mi}^{(t)}, q^{(i)}\varepsilon_{mi}^{(t)}  \right) \\
    \varepsilon_{mi}^{(t)} &= y_m^{(t)}-y^{(t)}_{mi}
\end{split}
\end{equation}
for a given dataset $\{(\mathbf{y}_m, \mathbf{x}_m, \mathbf{c}_m)\}_{m=1}^M$ and predictions $\{[\mathbf{y}_{mi}]_{i=1}^Q = [[y^{(t)}_{mi}]_{t=1}^T]_{i=1}^Q\}_{m=1}^M$. Note that this corresponds to the conventional quantile loss averaged over all output features and quantile levels. This setup can easily be converted to the day-ahead forecasting task at hand by replacing $\mathbf{y}_m$ with $\mathbf{x}_{un}$, $\mathbf{x}_{m}$ with $\mathbf{x}_{u(n-1)}$, and $\mathbf{c}_{m}$ with $\mathbf{c}_{un}$.

\subsection{Training}
We applied the same training, neural network, and constraint settings on GUIDE-VAE as in \cite{bolat2024guide}, including features like the number of layers and neurons, learning rate, early stopping, and learning rate schedule.
Also, we set the size of the quantile regression baseline model equal to the GUIDE-VAE in terms of number of learnable parameters to keep the comparison fair.

\subsection{Performance Metrics\protect\footnote{Starting from here, we assume that each data point $\mathbf{x}_{un}$ belongs to the test set, i.e. $n=0$ corresponds the first day of the testing set.}}

\subsubsection{Benchmarking Metrics}
These metrics are used to test the performance of GUIDE-VAE against the quantile regression model. Since the quantile regression model does not yield an explicit probability distribution against which to compare, we adapted GUIDE-VAE to the quantile forecasting setting. For this, we took $S$ samples from $p_{\psi,\mathbf{U}}(\mathbf{x}_{un}|\mathbf{c}_{un})$ for each ($u$,$n$)-pair as $\hat{\mathbf{x}}_{uns} \sim p_{\psi,\mathbf{U}}(\mathbf{x}_{un}|\hat{\mathbf{z}}_s,\mathbf{c}_{un}) 
, \forall s\in[S]$ where $\hat{\mathbf{z}}_s\sim p(\mathbf{z})$. Recall that each condition $\mathbf{c}_{un} = \left[\mathbf{x}_{u(n-1)}, \hat{\theta}_u, \tilde{\mathbf{c}}_{n} \right]$ represents the look-back window and the contextual information required for forecasting $\mathbf{x}_{un}$. Then, we calculated the quantile values of each feature $x_{un}^{(t)}$ for given quantile levels $\mathbf{q}=[q^{(i)}]$ using these $S$ samples as 
\begin{equation}
    x_{uni}^{(t)} \coloneq \inf \lbrace x\in \mathbb{R}: q^{(i)}\leq  \frac{1}{S}\sum_{s=1}^S \mathbf{1}(\hat{x}_{uns}^{(t)} \leq x) \rbrace
\end{equation}
where $\mathbf{1}(.)$ is the indicator function. This pipeline of quantile extraction is depicted in Fig. \ref{fig:representations}.

Having access to quantile values for both forecasting models, we enlist the benchmarking metrics and their definitions:
\begin{itemize}
    \item Quantile Loss: Defined in \eqref{eq:quantile_loss} and used with proper replacements as described before.
    \item Interval Score \cite{gneiting2007strictly}: We apply this score only to quantile level pairs (intervals) that are symmetric around 0.5. For conciseness, we assume that all the quantile levels in $\mathbf{q}$ satisfy this condition, i.e. $q^{(i)}+q^{(Q-i+1)}=1$ and $q^{(\frac{Q}{2})}\neq0.5$.\footnote{Even though we included the median in the quantile levels in the experiments, we did not include it to Interval and Interval-Coverage scores.}    
    Consequently, the Interval score is calculated as
    \begin{equation}
        \begin{split}
            {Interval} &= \frac{1}{UN_\text{test}T\frac{Q}{2}} \sum_{u,n,t,i}^{U,N_\text{test},T,\frac{Q}{2}} \Big (x_{uni}^{(t)}-x_{un(Q-i+1)}^{(t)} \\ &+ \frac{2}{I_i} \big (\max(0, \varepsilon_{unti}^{\text{upper}})+\max(0, \varepsilon_{unti}^{\text{lower}}) \big) \Big )
        \end{split}
    \end{equation}
    where $I_i=q^{(Q-i+1)}-q^{(i)}$, $\varepsilon_{unti}^{\text{upper}} =  x_{un}^{(t)} - x_{un(Q-i+1)}^{(t)}$ and $\varepsilon_{unti}^{\text{lower}} =  x_{uni}^{(t)} - x_{un}^{(t)}$.
        
    \item Interval-Coverage Score \cite{feldman2021improving}: Similar to the Interval score, this score is applied only to symmetric quantile levels:
        \begin{equation}
        \begin{split}
            {InterCover}_i &= \\
            \frac{1}{UN_\text{test}T} &\sum_{u,n,t}^{U,N_\text{test},T} \mathbf{1}\left(x_{un}^{(t)} 
     \in [x_{uni}^{(t)},x_{un(Q-i+1)}^{(t)}]\right) \\
            {InterCover} &= \frac{2}{Q} \sum_{i=1}^\frac{Q}{2} | {InterCover}_i - I_i |.
        \end{split}
        \end{equation}
\end{itemize}

\subsubsection{Ablation Metrics} Since the benchmarking metrics ignore the correlated structure of the forecasts of GUIDE-VAE by calculating the quantiles marginally, they do not reflect the added benefits of employing PDCC in GUIDE-VAE. Thus, we conducted an ablation study to investigate the effects of user embeddings and PDCC on the forecasting task by leveraging the explicit density modelling.\footnote{Recall that without these, GUIDE-VAE corresponds to a regular conditional VAE, which can also be used for the day-ahead forecasting.}
\begin{itemize}
    \item Average log-likelihood:\footnote{VAEs do not yield an exact distribution model. However, the log-likelihood value of a given data point can be approximated using importance sampling as explained in \cite{bolat2024guide}.} 
    \begin{equation}
        \text{ALL} = \frac{1}{UN_\text{test}}\sum_{u,n}^{U,N_\text{test}} \log p_{\psi,\mathbf{U}}(\mathbf{x}_{un}|\mathbf{c}_{un})
    \end{equation}
\end{itemize}

%% file: chapters/results.tex
\section{Results}
\input{chapters/results_table}

\begin{figure*}[ht]
    \centering
    \includegraphics[width=0.97\linewidth]{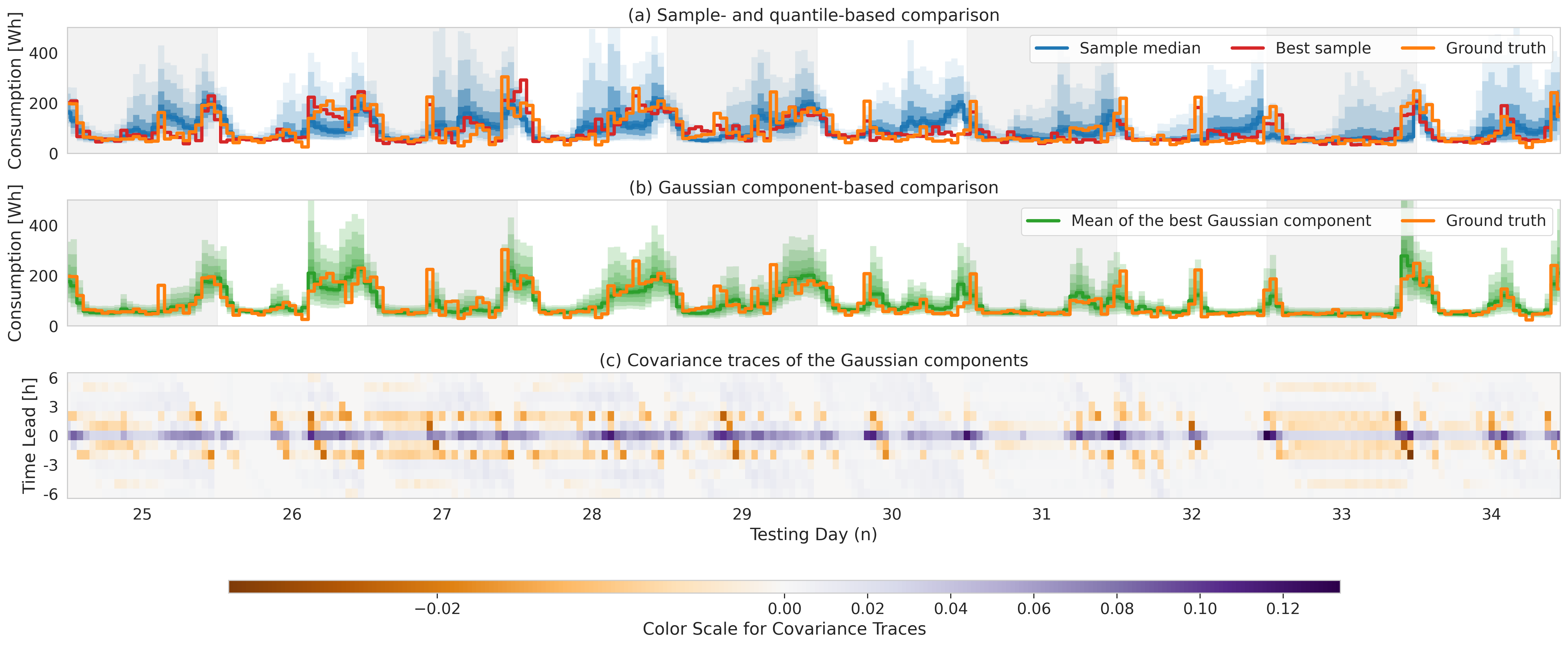}
    \caption{ Possible types of 10 consecutive day-ahead forecast sequences provided by GUIDE-VAE for $S=500$ samples. (a) The resulting quantile values derived from these samples (in blue), the ground truth (in orange) and, the closest trace to it out of 500 samples (in red). (b) The marginal trace of the best-fitting sampled Gaussian component out of 500 samples (in green). (c) Covariance trace of the components in (b), representing the temporal dependencies.}
    \label{fig:output}
\end{figure*}

We compared the performance of the GUIDE-VAE forecaster over the baseline model with different settings and tabulated the results in Table \ref{tbl:enter-label}. It can be seen that GUIDE-VAE outperforms the quantile regression neural network in all benchmarking metrics. This is particularly impressive in terms of the $QuantileLoss$ score since the baseline model is directly trained to minimize it while GUIDE-VAE is adapted to quantile forecasting ad-hoc. Another conclusion is that employing the user embeddings as contextual information elevates the performance of both models.

Besides the performance over the baseline model, the ablation study within GUIDE-VAE suggests that it is superior to a regular conditional VAE-based forecaster, which can be concluded by inspecting both the benchmarking and ablation metrics. We also see that the most significant performance improvement comes from employing the pattern dictionary.

We also visually showcase the forecasting capabilities of GUIDE-VAE in Fig. \ref{fig:output}. For this, we ran a sub-experiment where we chose a random user $\hat{u}$ and got their testing set $\mathcal{X}^\text{test}_{\hat{u}}$, which consists of consecutive daily profiles of the testing days. Then, we generated the parameters of the day-ahead (approximate) forecast distributions for each day as
\begin{equation*}
\begin{split}
    \{\hat{\mu}_{\hat{u}ns}\}_{s=1}^{S} &= \Big \{ f^\mu_\psi\big(\mathbf{z}_s,\mathbf{c}_{\hat{u}n} \big) \Big \}_{s=1}^{S} \\
    \{\hat{\Sigma}_{\hat{u}ns}\}_{s=1}^{S} &= \Big \{ \mathbf{U}\text{diag}\big ( f^{\tilde{\sigma}}_\psi (\mathbf{z}_s,\mathbf{c}_{\hat{u}n})\big)^2\mathbf{U}^\top + \xi\mathbf{I} \Big \}_{s=1}^{S}
\end{split}
\end{equation*}
where $\mathbf{c}_{\hat{u}n} = \left[\mathbf{x}_{\hat{u}(n-1)}, \hat{\theta}_{\hat{u}}, \tilde{\mathbf{c}}_{n} \right]$. These parameter collections correspond to the approximate forecasting pdf parameters in \eqref{eq:forecasting_pdf}, i.e. the Gaussian mixture components. Note that the forecasts of consecutive days are independent of each other.

Fig. \ref{fig:output}a depicts the ground truth consumption profile of the user $\hat{u}$, concatenated in time as $\mathbf{x}_{\hat{u}}^\top=[\mathbf{x}_{\hat{u}n}^\top]_{n=1}^{N_\text{test}}$. In the background, the quantiles derived from the samples $\hat{\mathbf{x}}_{\hat{u}ns} \sim \mathcal{N}\left(\mathbf{x}_{\hat{u}n}; \hat{\mu}_{\hat{u}ns}, \hat{\Sigma}_{\hat{u}ns}\right)$ are given. However, as stressed earlier, quantile traces do not convey the temporal dependencies that appear in point estimations. To showcase the ``realism" of the point estimates $\hat{\mathbf{x}}_{\hat{u}ns}$, we selected the best performing point estimates for each day as $\hat{\mathbf{x}}_{\hat{u}n}^* = \underset{\{\hat{\mathbf{x}}_{\hat{u}ns}\}_{s=1}^{500}}{\text{argmax}} ~\lVert  \mathbf{x}_{\hat{u}n} - \hat{\mathbf{x}}_{\hat{u}ns}\rVert_2$ and concatenated them in time to obtain the best prediction trace. As can be seen, even with a relatively low sample size of 500, GUIDE-VAE generated samples that capture the temporal dynamics very well. Also, note the discrepancy between the quantile traces and the best-performing trace, verifying that simplifying the full probabilistic forecast into quantiles underestimates the possible prediction capabilities of the forecaster, as described before in Fig. \ref{fig:representations}.

Another perspective on the forecast is obtained by considering directly its probabilistic information. 
With the full forecast being a mixture of Gaussians, we focus on the information contained in a single multivariate Gaussian distribution, which provides both interpretability and uncertainty quantification. 
To illustrate this in relation to our ground truth profile, 
we take $S=500$ samples, resulting in 500 multivariate Gaussian distributions,\footnote{Reminder: GUIDE-VAE works with two-level sampling. After the first level, we have a collection of sampled distributions, not points.} for each testing day and selected the best-performing MoG component for each day as\footnote{This is only for illustration since it is impossible to find the best-performing component without observing the event.}
\begin{equation*}
    \hat{\mu}^*_{\hat{u}n}, \hat{\Sigma}^*_{\hat{u}n} = \underset{\{(\hat{\mu}_{\hat{u}ns}, \hat{\Sigma}_{\hat{u}ns})\}_{s=1}^{500}}{\text{argmax}} \mathcal{N}(\mathbf{x}_{\hat{u}n};\hat{\mu}_{\hat{u}ns},\hat{\Sigma}_{\hat{u}ns}).
\end{equation*}
Then, we extracted the standard deviations as $\hat{\sigma}^*_{\hat{u}n} = \text{diag}(\hat{\Sigma}^*_{\hat{u}n})^{\frac{1}{2}}$ and placed them around each $\hat{\mu}^*_{\hat{u}n}$ with varying factors, i.e. $\hat{\mu}^*_{\hat{u}n} + \alpha \hat{\sigma}^*_{\hat{u}n}$ where $\alpha\in \{0.25, .5, 1, 1.5, 2\}$. Fig. \ref{fig:output}b depicts the traces after concatenation. Note the coverage of the prediction trace and how well it traces the ground truth.

Since visualizing only the standard deviations fails to represent the dependency structure embedded in $\hat{\Sigma}^*_{\hat{u}n}$, we extracted its ``covariance trace" by taking its diagonal band of size 13. This resulted in a matrix $G^*_{\hat{u}n} \in \mathbb{R}^{13\times24}$ where $G_{it}$ represents the covariance between the $t$-th and $(t-i+6)$-th time step. The resulting covariance trace after concatenation is given in Fig. \ref{fig:output}c. Even though it is visually challenging to interpret the covariance traces, we argue that they can be valuable in intra-day operations to foresee the error propagations in forecasts. On another note, the non-diagonal covariance structure of the multivariate Gaussian estimates allows the decision-maker to refine their predictions every hour by conditioning.

%% file: chapters/results_table.tex
\begin{table*}[ht!]
\centering
\caption{Quantitative Results on the Test Set}
\label{tbl:enter-label}
\begin{tblr}{
  cells = {c},
  cell{1}{1} = {r=2}{},
  cell{1}{2} = {r=2}{},
  cell{1}{3} = {r=2}{},
  cell{1}{4} = {c=3}{},
  cell{3}{1} = {r=4}{},
  cell{7}{1} = {r=2}{},
  hline{1,3,7,9} = {-}{},
  hline{2} = {4-7}{},
  hline{4-6,8} = {2-7}{},
}
\textbf{Model}                        & {\textbf{User Embedding }\\\textbf{Size~}($K$)} & {\textbf{Pattern Dictionary }\\\textbf{Size~}($V$)} & \textbf{Benchmarking Metrics } &                &                   & \textbf{Ablation Metrics} \\
                                      &                                                 &                                                     & $Quantile Loss$                & $Interval$     & $InterCover$      & $ALL$                     \\
GUIDE-VAE                             & 100                                             & 100                                                 & \textbf{56.11}                 & \textbf{955.1} & 37.99e-2          & \textbf{-2.163}           \\
                                      & 100                                             & 0                                                   & 57.34                          & 1008.6         & 36.55e-2          & -3.140                    \\
                                      & 0                                               & 100                                                 & 58.65                          & 993.2          & 37.88e-2          & -2.495                    \\
                                      & 0                                               & 0                                                   & 60.64                          & 1024.5         & \textbf{36.38e-2} & -3.569                    \\
{Quantile Regression\\Neural Network} & 100                                             & -                                                   & 58.43                          & 1029.0         & 40.01e-2          & -                         \\
                                      & 0                                               & -                                                   & 59.60                          & 1046.6         & 40.21e-2          & -                         
\end{tblr}
\end{table*}

%% file: chapters/conclusion.tex
\section{Conclusion}

This study introduced a probabilistic day-ahead forecasting framework based on GUIDE-VAE, a conditional VAE model designed for multi-entity datasets. GUIDE-VAE incorporates key features such as a pattern dictionary and entity embeddings, enabling efficient modelling of complex data structures. Unlike traditional approaches, GUIDE-VAE supports scalable, entity-specific probabilistic forecasting with a single model, eliminating the inefficiency of training separate models for individual entities. Its flexible outputs range from interpretable point estimates to full probability distributions that capture uncertainty and temporal dependencies.

Household electricity consumption was used as a challenging case study due to its highly stochastic nature and large number of entities. Our results demonstrate that GUIDE-VAE significantly outperforms conventional quantile regression techniques across key metrics. Additionally, the ablation study highlights the critical role of PDCC in capturing temporal correlations and improving forecast realism and accuracy across diverse settings.

The training phase is the most computationally demanding aspect of the GUIDE-VAE forecaster. It consists of two stages: entity embedding extraction and VAE training, where the latter is most demanding, and employing pattern dictionaries does not dramatically affect the VAE training time. In deployment scenarios, retraining the model, e.g., once a month, is not too burdensome. We note that the independent nature of the sampled profiles makes GUIDE-VAE's inference easily parallelizable, making it fast in practice.

In conclusion, GUIDE-VAE offers a powerful and generalizable framework for probabilistic forecasting tasks in energy systems and beyond. Its ability to handle multi-entity data, provide entity-specific forecasts, and deliver probabilistic outputs makes it a versatile tool for decision-makers aiming to improve risk management, resource allocation, and operational planning. Future work will focus on extending GUIDE-VAE to other multi-entity datasets, such as feeders and wind turbines, exploring the use of its covariance structures for refining intra-day forecasts and uncovering deeper insights into temporal dependencies.

%% file: main.bbl
\begin{thebibliography}{10}
\providecommand{\url}[1]{#1}
\csname url@samestyle\endcsname
\providecommand{\newblock}{\relax}
\providecommand{\bibinfo}[2]{#2}
\providecommand{\BIBentrySTDinterwordspacing}{\spaceskip=0pt\relax}
\providecommand{\BIBentryALTinterwordstretchfactor}{4}
\providecommand{\BIBentryALTinterwordspacing}{\spaceskip=\fontdimen2\font plus
\BIBentryALTinterwordstretchfactor\fontdimen3\font minus \fontdimen4\font\relax}
\providecommand{\BIBforeignlanguage}[2]{{%
\expandafter\ifx\csname l@#1\endcsname\relax
\typeout{** WARNING: IEEEtran.bst: No hyphenation pattern has been}%
\typeout{** loaded for the language `#1'. Using the pattern for}%
\typeout{** the default language instead.}%
\else
\language=\csname l@#1\endcsname
\fi
#2}}
\providecommand{\BIBdecl}{\relax}
\BIBdecl

\bibitem{wang2024generative}
X.~Wang, L.~Tong, and Q.~Zhao, ``Generative probabilistic time series forecasting and applications in grid operations,'' in \emph{2024 58th Annual Conference on Information Sciences and Systems (CISS)}.\hskip 1em plus 0.5em minus 0.4em\relax IEEE, 2024, pp. 1--6.

\bibitem{haben2021review}
S.~Haben, S.~Arora, G.~Giasemidis, M.~Voss, and D.~V. Greetham, ``Review of low voltage load forecasting: Methods, applications, and recommendations,'' \emph{Applied Energy}, vol. 304, p. 117798, 2021.

\bibitem{kingma2013auto}
D.~P. Kingma, ``Auto-encoding variational bayes,'' \emph{arXiv preprint arXiv:1312.6114}, 2013.

\bibitem{rezende2015variational}
D.~Rezende and S.~Mohamed, ``Variational inference with normalizing flows,'' in \emph{International conference on machine learning}.\hskip 1em plus 0.5em minus 0.4em\relax PMLR, 2015, pp. 1530--1538.

\bibitem{dumas2022deep}
J.~Dumas, A.~Wehenkel, D.~Lanaspeze, B.~Corn{\'e}lusse, and A.~Sutera, ``A deep generative model for probabilistic energy forecasting in power systems: normalizing flows,'' \emph{Applied Energy}, vol. 305, p. 117871, 2022.

\bibitem{bolat2024guide}
K.~B{\"o}lat and S.~Tindemans, ``{GUIDE-VAE}: Advancing data generation with user information and pattern dictionaries,'' \emph{arXiv preprint arXiv:2411.03936}, 2024.

\bibitem{jrhilifa2024forecasting}
I.~Jrhilifa, H.~Ouadi, A.~Jilbab, and N.~Mounir, ``Forecasting smart home electricity consumption using vmd-bi-gru,'' \emph{Energy Efficiency}, vol.~17, no.~4, p.~35, 2024.

\bibitem{chen2022constructing}
X.~Chen, C.~Zanocco, J.~Flora, and R.~Rajagopal, ``Constructing dynamic residential energy lifestyles using latent dirichlet allocation,'' \emph{Applied Energy}, vol. 318, p. 119109, 2022.

\bibitem{quesada2024electricity}
C.~Quesada, L.~Astigarraga, C.~Merveille, and C.~E. Borges, ``An electricity smart meter dataset of spanish households: insights into consumption patterns,'' \emph{Scientific Data}, vol.~11, no.~1, p.~59, 2024.

\bibitem{wang2022contextual}
C.~Wang, S.~H. Tindemans, and P.~Palensky, ``Generating contextual load profiles using a conditional variational autoencoder,'' in \emph{2022 IEEE PES Innovative Smart Grid Technologies Conference Europe (ISGT-Europe)}, 2022, pp. 1--6.

\bibitem{gneiting2007strictly}
T.~Gneiting and A.~E. Raftery, ``Strictly proper scoring rules, prediction, and estimation,'' \emph{Journal of the American statistical Association}, vol. 102, no. 477, pp. 359--378, 2007.

\bibitem{feldman2021improving}
S.~Feldman, S.~Bates, and Y.~Romano, ``Improving conditional coverage via orthogonal quantile regression,'' \emph{Advances in neural information processing systems}, vol.~34, pp. 2060--2071, 2021.

\end{thebibliography}
